\documentclass[letterpaper, 10pt, conference]{ieeeconf}
\IEEEoverridecommandlockouts
\usepackage{flushend}
\usepackage{marginnote}
\usepackage{graphics}
\usepackage[pdftex]{graphicx}
\DeclareGraphicsExtensions{.pdf,.png,.jpg}
\usepackage[font={small}]{caption}
\usepackage{subcaption}
\usepackage[rightcaption]{sidecap}
\usepackage{pbox}
\usepackage{mathtools}
\usepackage{amsmath, amssymb, amscd}
\usepackage{ wasysym } %
\usepackage{amsfonts}
\usepackage{mathptmx} %
\usepackage{gensymb} 
\DeclareMathAlphabet{\mathcal}{OMS}{lmsy}{m}{n}
\DeclareSymbolFont{largesymbols}{OMX}{cmex}{m}{n}
\usepackage{textcomp} %
\usepackage[ruled,vlined,linesnumbered]{algorithm2e} 
\usepackage{algorithmicx, algpseudocode} 
\usepackage{array} %
\usepackage{tabularx}
\usepackage{multirow}
\usepackage{multicol}
\usepackage{booktabs}
\usepackage[T1]{fontenc} 
\usepackage[utf8]{inputenc}
\usepackage[english]{babel} %
\usepackage{units}
\usepackage{bm}
\usepackage{times} %
\usepackage{xspace}
\usepackage{flushend}%
\usepackage{csquotes}
\usepackage{makeidx}
\usepackage{soul} %
\usepackage{subfiles} %
\let\labelindent\relax
\usepackage{enumitem}

\usepackage[protrusion=true,expansion=true]{microtype}
\usepackage[yyyymmdd]{datetime}

\date{\protect\formatdate{1}{1}{2001}}

\usepackage{url}
\makeatletter
\g@addto@macro{\UrlBreaks}{\UrlOrds}
\makeatother
\usepackage{color}
\usepackage[usenames,dvipsnames,table]{xcolor}
\usepackage[pdfborder={0 0 0.5}]{hyperref}
\hypersetup{
    colorlinks=true,
    linkcolor=black,
    citecolor=black,
    filecolor=cyan,
    urlcolor=black
}

\usepackage{amsmath}
\usepackage{balance}
\usepackage{caption}
\numberwithin{equation}{section}

\begin{document}

\title{\LARGE \bf
StRETcH: a Soft to Resistive Elastic Tactile Hand
}

\author{Carolyn Matl, Josephine Koe, Ruzena Bajcsy
\thanks{All authors are affiliated with the Department of Electrical Engineering and Computer Science, University of California, Berkeley, CA, USA}
}

\maketitle

\begin{abstract}
Soft optical tactile sensors enable robots to manipulate deformable objects by capturing important features such as high-resolution contact geometry and estimations of object compliance. This work presents a variable stiffness soft tactile end-effector called StRETcH, a Soft to Resistive Elastic Tactile Hand, that is easily manufactured and integrated with a robotic arm. An elastic membrane is suspended between two robotic fingers, and a depth sensor capturing the deformations of the elastic membrane enables sub-millimeter accurate estimates of contact geometries. The parallel-jaw gripper varies the stiffness of the membrane by uniaxially stretching it, which controllably modulates StRETcH's effective modulus from approximately 4kPa to 9kPa. This work uses StRETcH to reconstruct the contact geometry of rigid and deformable objects, estimate the stiffness of four balloons filled with different substances, and manipulate dough into a desired shape.
\end{abstract}


\IEEEpeerreviewmaketitle

\section{Introduction}

With the advent of home robotics, robotic interaction with deformable objects such as electrical cables, bed sheets, and pizza dough is inevitable \cite{sanchez2018robotic}. 
In particular, effective dexterous manipulation of solid deformable objects like dough requires the estimation of both contact geometry \cite{yousef2011tactile} as well as compliance \cite{luo2017robotic}.
This work aims to address the problem of estimating both stiffness and geometry while a robot is in contact with a solid deformable object.

To achieve this, this work proposes the development and use of a Soft to Resistive Elastic Tactile Hand, StRETcH. Different tactile sensors enable a robot to make various important measurements while in contact with an object, such as contact area, force, texture, and slip detection. Within the diverse set of existing tactile sensors, soft tactile sensors are particularly well-suited for interacting with deformable or fragile objects due to their inherent compliance. Prior work has demonstrated the clear benefits of soft material sensors, including sensitivity to stiffness for tumor detection \cite{ward2018tactip}, high spatial resolution for classifying different textures \cite{yuan2018active}, and handling of delicate objects like wine glasses \cite{kuppuswamy2020soft}. The above mentioned soft tactile sensors are all optically-based, which enables high resolution contact imaging without the need to embed rigid components in the soft interface.

StRETcH is not only a soft optically-based tactile hand, but it is also able to adjust its stiffness and therefore vary the force it exerts on the environment. This work draws on our prior work where a soft dome-shaped fingertip called SOFTcell was developed, which explored the idea of a controllable-stiffness tactile sensor through pneumatic actuation \cite{mcinroe2018towards}. An improved design of SOFTcell, which uses a depth sensor to image the deforming membrane \cite{huang2019depth}, was later applied as an assistive robotic interface to dynamically support a human arm \cite{huang2020high} as well as to perform geometry-dependent tasks \cite{huang2020robot}. In this work, we modify our prior sensor by designing the membrane to be planar rather than hemispherical, and varying its stiffness using a robotic gripper to mechanically stretch the membrane uniaxially instead of increasing its stiffness through pneumatic actuation. The benefits of this new design include the simple manufacturing process of the end-effector and easier control of the stiffness of the membrane via mechanical stretching rather than pneumatic actuation. This work aims to utilize the variable stiffness feature of the end-effector to estimate stiffnesses of different deformable objects as well as to modulate the force exerted on a solid deformable object in a manipulation task. 

The main contributions of this paper are:
\begin{itemize}
    \item The Soft to Resistive Elastic Tactile Hand (StRETcH), a new variable stiffness tactile end-effector that is easy to manufacture, integrate, and use with a robotic arm. 
    \item Modeling and experimental characterization of StRETcH and its sensitivity to contact area and object stiffness.
    \item A robotic demonstration using StRETcH to actively perceive and shape a piece of soft dough. 
\end{itemize}

\begin{figure}[t]
	\includegraphics[width=\linewidth]{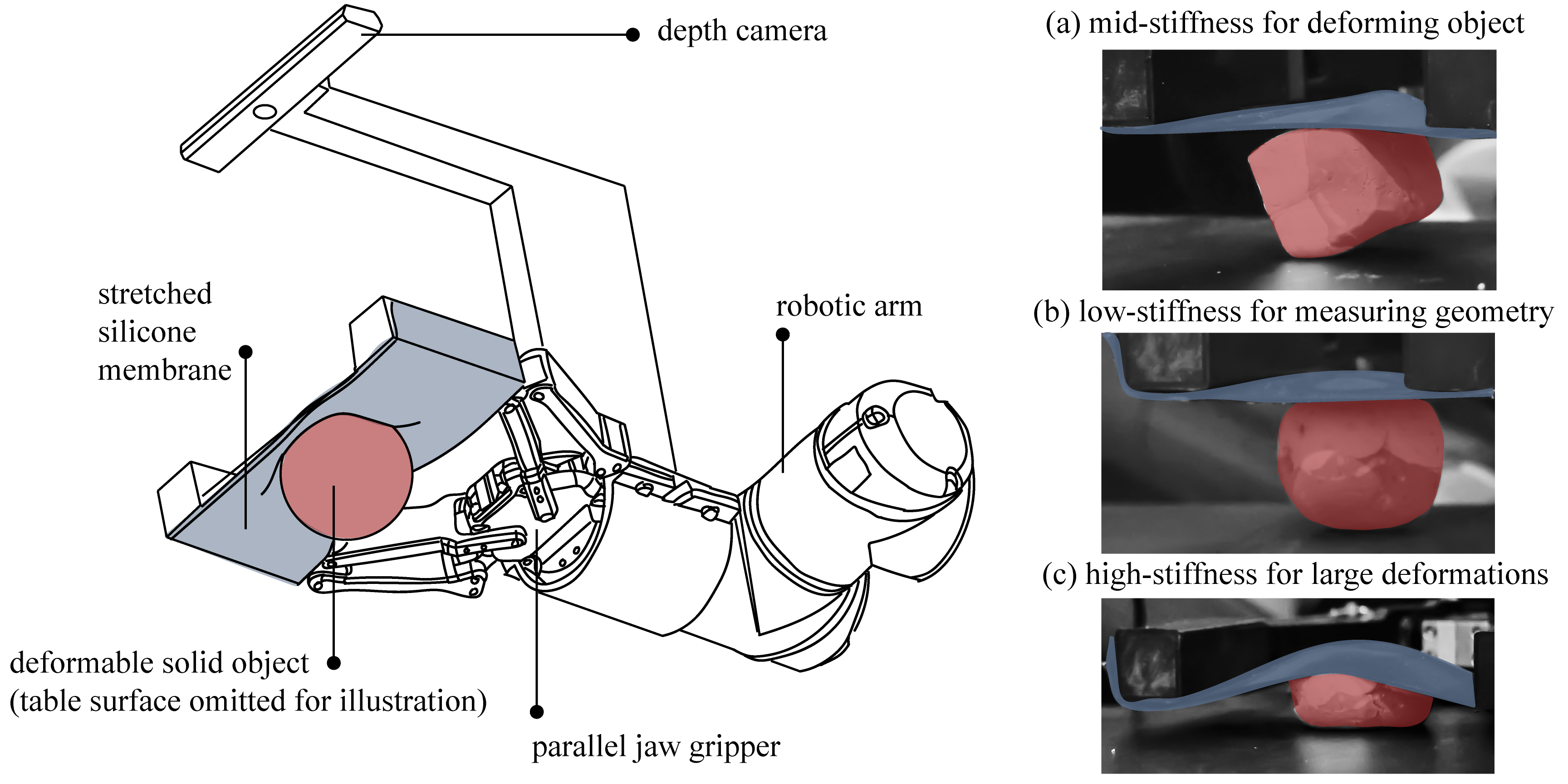}
	\centering
	\caption{(Left): An orthographic view of StRETcH, mounted on a robotic arm. An elastic membrane (blue) is attached to the robotic gripper, which modulates its stiffness through stretching when interacting with an object (red). (Right): (a) Mid-range stiffness is used to deform a soft object; (b) Low-range stiffness is used when measuring the object's geometry; (c) High-range stiffness is used to exert large forces on the object.}
	\label{fig:figure1}
	\vspace{-0pt}
\end{figure}

\section{Related Work}

\subsection{Soft Tactile Sensing}

Existing soft tactile sensors can be categorized by their underlying sensor technology -- mainly, resistive, capacitive, piezoelectric, and optical sensing \cite{yousef2011tactile}. Many resistive-based sensors rely on intricate composition of individual nodes to achieve high-dimensional contact imaging \cite{hammond2012soft, lee2019neuro, suh2014soft, drimus2011classification}. However, embedding rigid materials such as conductive wires within a soft skin limits the material's strain and therefore range of deformation. Thus, highly compliant fabric-like tactile materials \cite{buscher2015flexible, atalay2017batch, yoshikai2009development} and ultra-stretchable capacitive sensors \cite{larson2016highly, pu2017ultrastretchable} have been developed. These materials are largely useful for proprioception of soft robots \cite{larson2016highly} or soft exoskeletons \cite{buscher2015flexible, atalay2017batch}, but their low-dimensional signals are insufficient for relaying contact area information. While it has been shown that richer contact information can be extracted from much lower-dimensional tactile signals \cite{narang2020interpreting}, achieving this requires sophisticated modeling of the sensor as well as an extensive dataset relating raw signals to known geometric values. 

Recent developments in optical image-based soft tactile sensors decouple the soft interface from the sensing mechanism, thereby achieving both high deformability and high contact resolution. Examples of such sensors include FingerVision \cite{yamaguchi2016combining}, GelSight \cite{yuan2017shape}, TacTip \cite{ward2018tactip}, SOFTcell \cite{mcinroe2018towards}, and Soft-bubble \cite{alspach2019soft}. In particular, the SOFTcell design incorporated both optical sensing and pneumatic actuation. Specifically, SOFTcell consisted of an elastic hemispherical membrane, which was imaged from within by an RGB camera and modulated in stiffness via pneumatic pressurization. This design enabled the sensor to be used as a controllable stiffness tactile device and explored the inherent sensing-action duality. Similar designs that use depth sensing to image the membrane were later developed in \cite{huang2019depth} and \cite{alspach2019soft}. This work modifies the design of SOFTcell \cite{mcinroe2018towards} for ease of manufacture, integration, and use with a robotic arm. We aim to use its controllable variable stiffness to enable high resolution contact imaging as well as measure deformable object stiffness.

\subsection{Tactile Perception of Deformable Objects}

One particular advantage we would like to exploit with the variable-stiffness image-based hand design is that it allows for high-resolution contact geometry and stiffness estimation of a wide range of deformable objects. Conventional tactile sensors often enable measurements for one or the other, but rarely both. This is because stiffness estimates are typically made by measuring the displacement of a probe upon application of a known force \cite{shikida2003active, yussof2008determination}, active palpation \cite{drimus2011classification}, or via piezoelectric resonance \cite{ju2018variable, omata2004real}, which inherently are low-dimensional signals and cannot replicate the rich geometric information available in image-based sensors. Meanwhile, soft optical tactile sensors, while able to image high-dimensional contacts, are typically only capable of sensing a small range of stiffnesses before deforming the object itself. For example, a soft optical tactile sensor called GelSight was used to estimate the hardness of several objects via active palpation \cite{yuan2017shape}. However, the authors note that estimates of hardness are more accurate when the object hardness is close to the hardness of that of the sensor. Because the GelSight is a gel-filled tactile sensor, its elastic membrane has fixed stiffness and thus its sensitivity is limited to a narrower range of deformable objects. This work aims to address this limitation by enabling the elastic interface to vary in stiffness. 

\subsection{Manipulation of Deformable Objects}

This work makes use of StRETcH's variable-stiffness to manipulate deformable objects into a desired shape. Deformable object manipulation can be categorized into the manipulation of deformable linear objects, planar or cloth-like objects, and solid objects \cite{sanchez2018robotic}. Examples of linear object manipulation include rope manipulation \cite{yan2020learning} and knot tying \cite{sundaresan2020learning}, while planar object manipulation includes the folding of paper \cite{balkcom2008robotic} or cloth \cite{stria2014garment, miller2012geometric}.

StRETcH was designed for the manipulation of solid deformable materials such as clay or foam. 
Prior work in solid deformable object manipulation depended primarily on image-based feedback. For example, in \cite{navarro2016automatic, navarro2017fourier}, a bilateral manipulator uses features extracted from a stereo-camera and an estimation of deformation properties to shape unstructured soft objects like foam into desired two-dimensional shapes. The authors in \cite{shah2019morphing} take a sensor-less approach by designing a unique morphing skin that wraps around sculptable materials like clay to deform the material. In \cite{figueroa2016learning}, the authors pair a depth sensor with a force sensor attached to a rigid roller to flatten pizza dough into a circle.  StRETcH incorporates a soft morphing skin as in \cite{shah2019morphing} while maintaining a sense of contact force, as shown in \cite{figueroa2016learning}. When deforming solid materials, tactile sensing plays a large role in determining the shape and hardness of an object, making StRETcH appropriate for such applications.

\section{System Design}

\begin{figure}[t]
	\includegraphics[width=\linewidth]{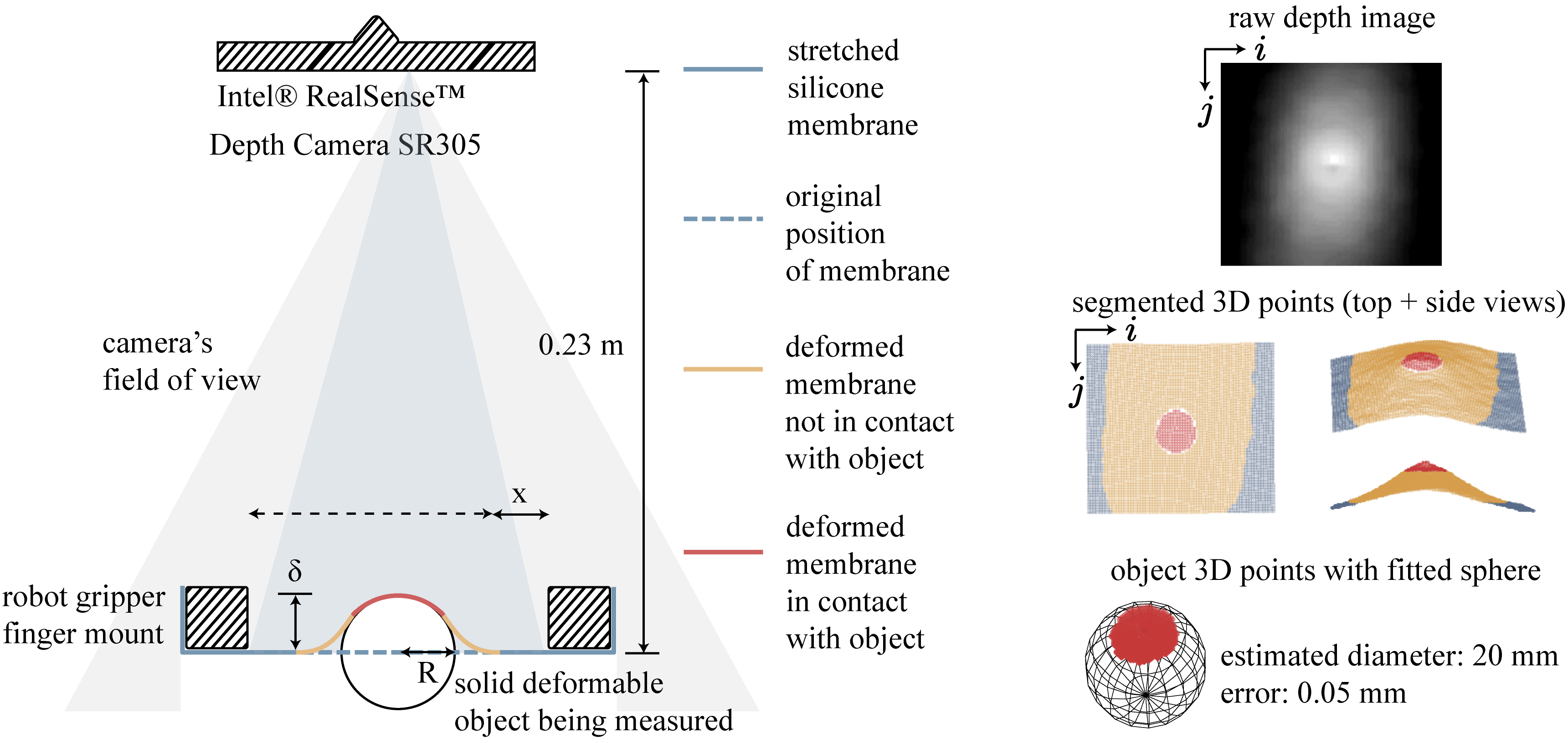}
	\centering
	\caption{(Left): A cross-sectional schematic of StRETcH. The membrane is stretched by $x$ mm. When the membrane makes contact with an object, it is deformed from its original position (dashed blue) and can be partitioned into undeformed (solid blue), deformed but not in contact (yellow), and deformed and in contact (red) segments. (Right): 3D points corresponding to the red, yellow, and blue segments are extracted from the depth image, resulting in a low error in estimated contact geometry.}
	\label{fig:setup}
	\vspace{-0pt}
\end{figure}

\subsection{Design and Fabrication}
StRETcH is comprised of an Intel RealSense Depth Camera SR305 and a flat silicone membrane attached to a Robotiq 2-Finger Adaptive Robot Gripper, which are both fixed to a Universal Robot UR5 robotic arm (see Figure \ref{fig:figure1}). The parallel-jaw gripper serves as the stretching mechanism for the membrane as it varies its gripper throw and can stretch the membrane up to $x=80$mm past its original length (see Figure \ref{fig:setup}). The silicone membrane was manufactured by pouring platinum-catalyzed EcoFlex 50
Silicone into a flat mold, which produced a 100mm $\times$ 90mm $\times$ 2mm rectangular membrane. Velcro was sewn onto strips of cloth that were embedded into the sides of the silicone during the curing stage. The silicone membrane was attached to 3D-printed gripper finger mounts with velcro.

\subsection{Contact Estimation}
\label{sec:contactestimation}

To interpret the deformations of the membrane captured by the depth sensor, we must first understand that the membrane surface can be categorically divided by its deformation state. Specifically, we follow the classification defined in \cite{huang2019depth} to segment the membrane into the following categories based on the estimated curvature at each point:
\begin{enumerate}[label=(\alph*)]
    \item Undeformed: the membrane lies flat at its initial depth and is assumed to not be in contact with the object.
    \item Deformed but not in contact: the membrane is deformed but not convex along either axis of the 2D depth image.
    \item Deformed and in contact: the membrane is deformed and is convex in either axis of the depth image.
\end{enumerate}

The segmentation of category (c) points enables estimation of contact area and geometry, as shown in red in Figure \ref{fig:setup}. Segmentation is based off of the proposed algorithm in \cite{huang2019depth}, which assumes that the object in contact with the membrane is flat or convex, and the contact area of the object is larger than a single surface element of the membrane. 

Under these assumptions, membrane segmentation begins by first isolating the membrane in the depth image. This is done by using a box mask that depends on the width of the gripper throw. The depth image points corresponding to the membrane are projected into 3D and transformed into coordinates relative to the membrane plane when it is flat and undeformed. Normal vectors are then estimated from the 3D point cloud of the membrane. These normal vectors and points are used to calculate curvature at each point. Let depth image pixel $(i, j)$ have 3D coordinates $\vec{p}(i, j)$ and normal vector $\vec{n}(i, j)$. Local curvature signals $\mathcal{K}_i$ and $\mathcal{K}_j$ along the $i$ and $j$ directions, respectively, can then be estimated at each depth image pixel using the method presented in \cite{nascimento2012brand}:
\begin{equation}
\begin{aligned}
\mathcal{K}_i(i, j) = \langle \vec{p}(i + 1, j) - \vec{p}(i - 1, j), \vec{n}(i + 1, j) - \vec{n}(i - 1, j) \rangle\\
\mathcal{K}_j(i, j) = \langle \vec{p}(i, j + 1) - \vec{p}(i, j - 1), \vec{n}(i, j + 1) - \vec{n}(i, j - 1) \rangle\\  
\end{aligned}
\end{equation}
If a curvature signal is positive, the point is locally convex in that direction; if it is negative, it is locally concave. Thus, 3D points are categorized as \textit{in contact} with the object if they exhibit sufficiently high curvature in either direction. Additionally, any points lying above the plane of these points are also considered to be in contact with the object, thus accounting for contacts with flat-faced objects.

\section{Model}
\label{sec:model}
The StRETcH membrane becomes more rigid as it is stretched. Characterization experiments show that the mechanical response of StRETcH is a function of the internal strain state of the elastic membrane, caused by uniaxial stretching from the gripper and transverse loading and displacement of the membrane from the object in contact, which depends on the contact geometry and depth of indentation.

The mechanics of Neo-Hookean membranes are complex and nonlinear, and, to the best of our knowledge, there does not exist an explicit model that calculates the mechanical response of a uniaxially stretched Neo-Hookean membrane under contact with elastic objects of varying geometries. There exist models for spherical \cite{bhatia1968finite, yang1971indentation, begley2004spherical} and cylindrical \cite{pamplona2014analytical, liu2018puncture} rigid indentation of stretched membranes, although these membranes are omnidirectionally stretched and our experimental data does not match their power laws. Recently, numerical simulations present an alternative to explicit models to, for example, calculate the strain due to uniaxial stress on rectangular sheets of rubber \cite{taylor2019finite, taylor2019finite}. However, numerical simulation is often too slow for real-time robotic use. 

Instead, we borrow and revise from prior work \cite{mcinroe2018towards, scott2004indentation} the idea of modeling the membrane as an equivalent linearly elastic solid half-space subject to the same contact conditions. This model is not meant to represent the internal strains of the \textit{membrane}, but rather, we adopt it as a template to express StRETcH's varying stiffness. We imagine that this equivalent half-space has an associated elastic modulus, which we refer to as the effective modulus E$^*$, to approximate the mechanical response of the membrane at its different stretch states.

Hertz contact theory \cite{johnson1987contact} relates the load force $F$ and indentation depth $\delta$ for frictionless contact between:
\begin{enumerate}
    \item an elastic half-space and an elastic sphere:
    \begin{equation}
    \label{eq:sphericalcontact}
        F = \frac{4}{3}(\delta^3 R)^{\nicefrac{1}{2}} (\frac{1-v^2}{E^*} + \frac{1-v_o^2}{E_o})^{-1}
    \end{equation}
    \item an elastic half-space and a rigid cylinder
    \begin{equation}
        F = 2R\delta E^*
    \end{equation}
    \item an elastic half-space and a rigid cone
    \begin{equation}
        F = \frac{2\delta^2}{\pi (1-v^2) \tan(\theta))}E^*
    \end{equation}
\end{enumerate}
where R is the radius of the sphere and cylinder, $\theta$ is the angle between the conical surface and the elastic surface, and $E^*$ and $v$ are the effective elastic modulus and Poisson's ratio of the half-space. We use the Poisson's ratio of ideal rubber (0.5) for $v$. $E_o$ and $v_o$ are the elastic modulus and Poisson's ratio of the elastic sphere, but in characterization experiments, the indenters are rigid, so it is assumed that $E^* << E_o$. Characterization experiments measure the mechanical response of the membrane against all three basic geometries and in various stretch states and indentation depths. All contact loads are centered, transverse, and normal to the membrane surface. Despite mechanical differences between elastic membranes and half-spaces, this simplified representation enables fast and empirically accurate load force estimations (Section \ref{sec:estimatingloadforce}).

\section{Experimental Characterization}

\subsection{Variable Stiffness}
\begin{figure}[t]
	\includegraphics[width=\linewidth]{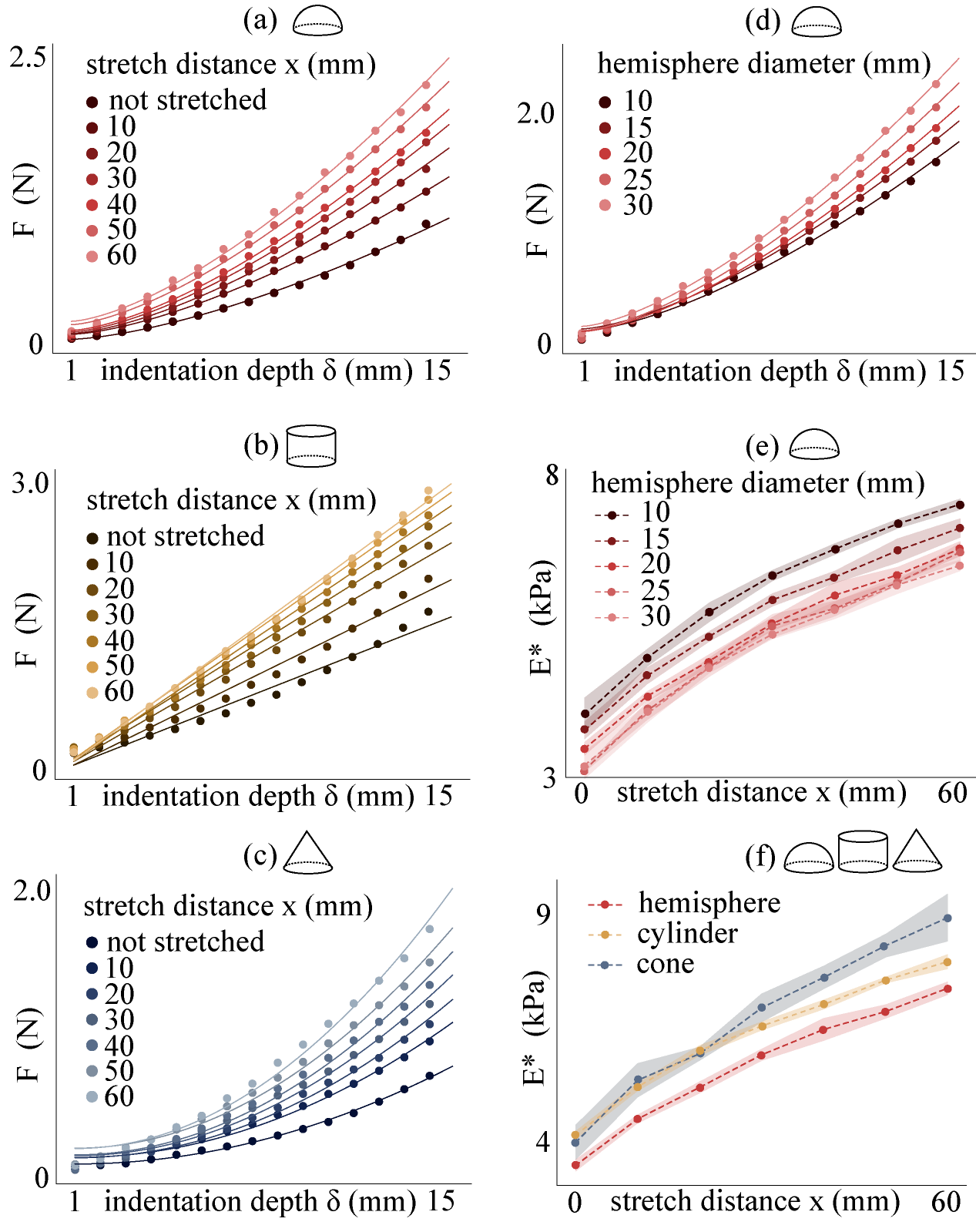}
	\centering
	\caption{(a) Indentation depth $\delta$ vs. load force $F$ for varying stretch states for a 20mm diameter hemispherical rigid indenter; (b) $\delta$ vs. $F$ for varying stretch states for a 20mm diameter cylindrical rigid indenter; (c) $\delta$ vs. $F$ for varying stretch states for a $45\deg$ conical rigid indenter; (d) $\delta$ vs. $F$ for varying diameter hemispherical indenters with a membrane stretched by 30mm; (e) Stretch distance $x$ vs. effective modulus $E^*$ for varying diameter hemispherical indenters; (f) $x$ vs. $E^*$ for indenters of varying geometry.}
	\label{fig:sensor_characterization}
	\vspace{-10pt}
\end{figure}
To characterize the varying stiffness of StRETcH under different contact conditions, we measured the contact force of a rigid hemisphere, cylinder, and cone, as well as rigid hemispheres with different diameters, on the membrane at various uniaxial stretch states ($x$) and indentation depths ($\delta$). A clamped ATI Axia80 EtherNet Force/Torque (F/T) sensor was used for force measurements, and 3D-printed probes of hemispheres, cylinders, and a cone were attached to the F/T sensor through friction coupling with a 3D-printed interface. The UR5 robot was commanded to press the membrane into the probe up to 15mm in depth in the z-axis of the F/T sensor, with 1mm steps, generating a vector of indentation depth to force data pairs. This was repeated 5 times each for 7 total stretch states, from zero applied tension to a Cauchy strain of 0.8 (the membrane was stretched up to 60mm beyond its original length). The dots in Figure \ref{fig:sensor_characterization} correspond to the force data measured at each contact condition. 

Using equations from Section IV, an effective modulus ($E^*$) was fit for each stretch state to the pairs of force to indentation depth data. The calibrated model with the estimated effective modulus is shown as the solid lines in Figure \ref{fig:sensor_characterization}. As shown in graphs (a)-(d), the resulting model adheres closely to the real data. Deviations occur for small indentations, but past 3mm indentations, the model performs well, especially under the condition of spherical and conical contact. Additionally, Figure \ref{fig:sensor_characterization}(d) shows that as the contact area increases, the load force increases as well. Finally, the averaged estimated effective moduli $E^*$ (in kPa) at each stretch state for the different scenarios (contact area and geometry) were plotted in Figure \ref{fig:sensor_characterization} (e) and (f), respectively. As shown in both graphs, the effective moduli monotonically increase in the operating region, which demonstrates that StRETcH indeed does have variable stiffness. The cylindrical and hemispherical data also show low variance over 5 trials, demonstrating that this variable stiffness is reasonably repeatable and controllable.
\subsection{Contact Estimation Sensitivity}
\label{sec:contactsensitivity}
\begin{figure}[t]
	\includegraphics[width=\linewidth]{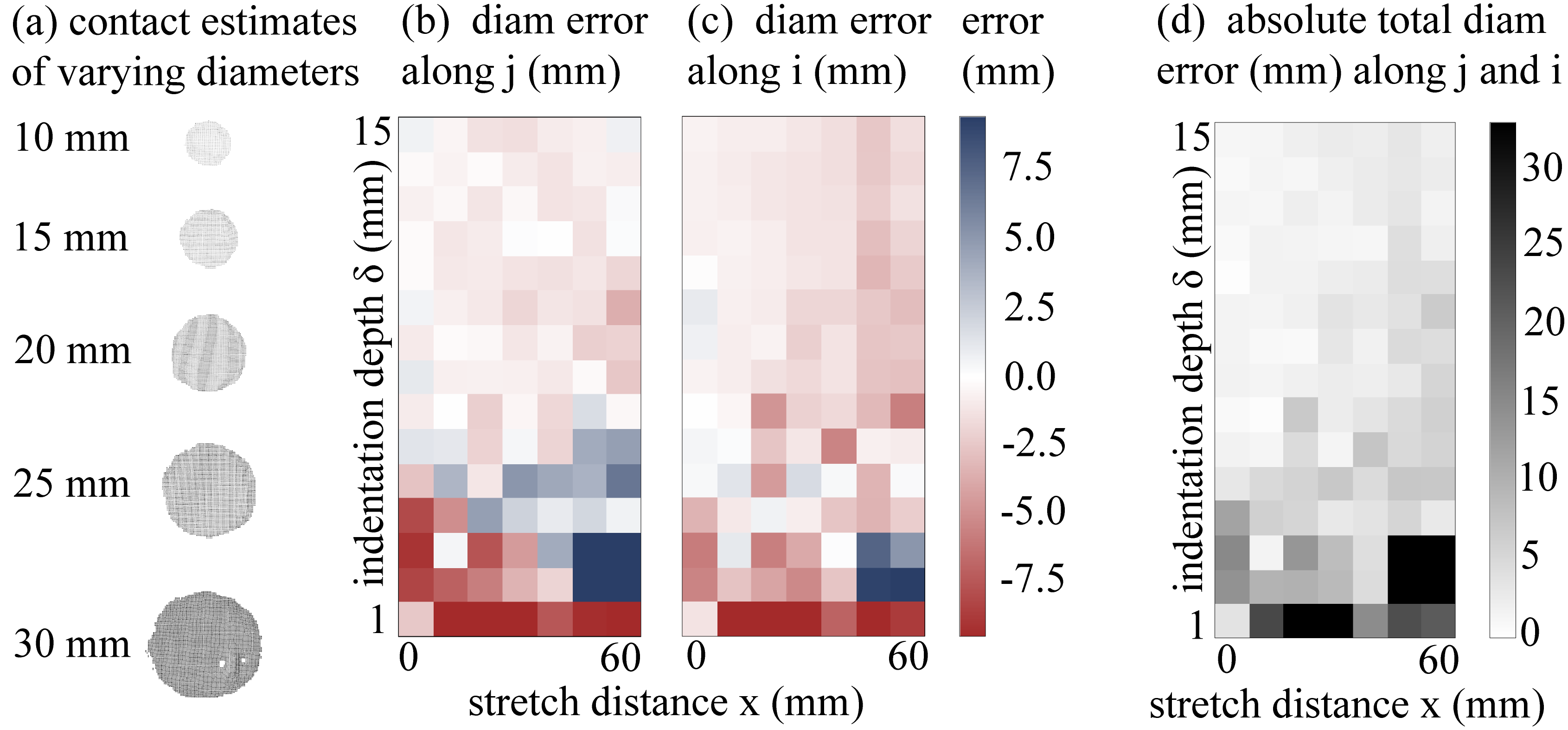}
	\centering
	\caption{(a) Example estimated contacts with cylinders of varying diameters; (b) Errors in estimated diameter along the stretch direction $j$ for different indentation depths $\delta$ and stretch distances $x$; (c) Errors in estimated diameter in the $i$ direction for different $\delta$ and $x$; (d) Absolute total error in estimated diameter along the $i$\&$j$ directions for different $\delta$ and $x$. (b-d) demonstrate higher contact estimation accuracy for larger $\delta$ and lower $x$.}
	\label{fig:sensor_sensitivity}
	\vspace{-0pt}
\end{figure}
In addition to characterizing StRETcH's variable stiffness, membrane indentation experiments were also performed to quantify the sensitivity of StRETcH in measuring contact geometry (see Section \ref{sec:contactestimation}). Five right circular cylinder probes were used in this characterization, ranging from 10mm to 30mm in diameter. For each cylinder, the UR5 was again commanded to press the membrane into the probe up to 15mm, in the direction of the length of the cylinder, for the 7 stretch conditions characterized in the prior section. The error in estimated diameter along the $i$ and $j$ directions were calculated for each scenario and depicted in Figure \ref{fig:sensor_sensitivity}. 

As shown in Figure \ref{fig:sensor_sensitivity}(a), the sensor is capable of sub-millimeter accuracy when estimating contact of the varying circular faces. However, this level of sensitivity arose under specific contact conditions. Figures \ref{fig:sensor_sensitivity}(b) and (c) visualize the diameter error for the 20mm cylinder under different indentation depths and stretch distances in the $i$ and $j$ directions. As shown in these two plots, in general, the high-sensitivity region corresponds to larger indentation depths and lower stretch states.  The high-accuracy region improved with decreasing diameter, since larger flat contact areas resulted in lower curvatures registered by the contact estimation algorithm. Indentation depths of less than 2mm tended to perform poorly for all five cylinders.  Furthermore, it appears that the more the membrane is uniaxially stretched, the greater the indentation depth is needed to achieve sub-millimeter accuracy. This matches the intuition that the stiffer the membrane is, the greater the force necessary for the membrane to conform to the geometry. 

\subsection{Estimating Load Force from Estimated Contact}
\label{sec:estimatingloadforce}

\begin{figure}[t]
	\includegraphics[width=\linewidth]{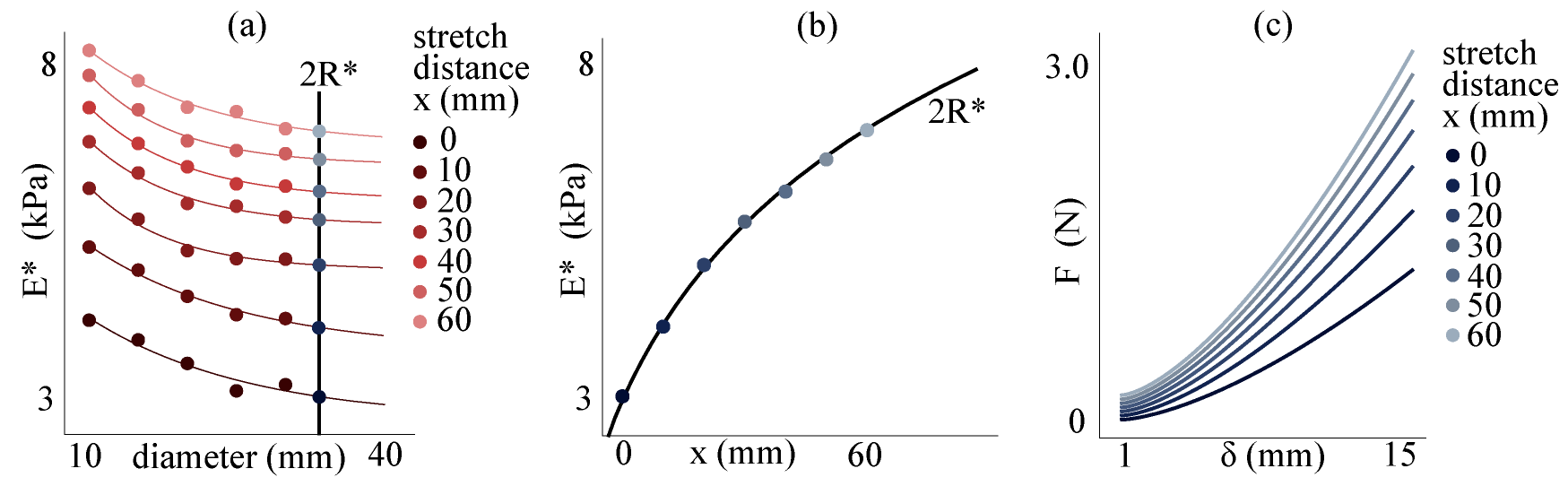}
	\centering
	\caption{(a) Alternative visualization of Figure \ref{fig:sensor_characterization}(e). Negative exponential curves are fitted to the datapoints (which are the average $E^*$ over five trials). The vertical line denotes corresponding effective moduli for an estimated spherical contact with radius $R^*$. (b) A logarithmic function is fit to the corresponding points lying on the vertical line $2R^*$ in plot (a). The fitted function estimates $E^*$ given the stretch distance $x$ for the particular geometry of radius $R^*$. (c) Generated $\delta$ vs. $F$ functions for the specific radius $R^*$. }
	\label{fig:loadforce_estimation}
	\vspace{-0pt}
\end{figure}

The above characterization experiments present StRETcH as a multimodal system: at low stretch states, the hand has high sensitivity to contact geometry, and higher stretch states enable greater contact forces. Here, we examine the accuracy of load force estimation given the estimated contact geometry at a particular indentation depth $\delta$ and stretch distance $x$. 

Let us consider contact with a rigid spherical object that has an estimated radius of $R^*$. To estimate the contact force of StRETcH with the object, we need to derive a model of $E^*(x)$ for the specific radius $R^*$.
First, note that the datapoints in Figure \ref{fig:sensor_characterization}(e) can be equivalently represented as in Figure \ref{fig:loadforce_estimation}(a). For each stretch distance, a negative exponential curve is fit, where each curve is a function of the contact diameter. This enables estimations of $E^*$ for a particular contact radius $R^*$ at different stretch distances. A logarithmic function can then be fit (see Figure \ref{fig:loadforce_estimation}(b)) to model $E^*$ for different stretch distances $x$, given the estimated radius $R^*$. Load force $F^*$ is thus estimated by plugging $E^*$, $R^*$ and $\delta$ into Equation \ref{eq:sphericalcontact}. Figure \ref{fig:loadforce_estimation}(c) shows $\delta-F$ curves for $R^*$ and various values of $x$, which are generated using the model in Figure \ref{fig:loadforce_estimation}(b).

We tested the accuracy of estimating $F^*$ given the estimated radius $R^*$ with leave-one-out cross-validation. Specifically, using four of the five curves in Figure \ref{fig:sensor_characterization}(e) to fit the model for $E^*$ given $x$ and $R^*$, we tested the accuracy of the estimated $F^*$ against ground truth for indentations of the left-out data. On average, the estimated load force had an error of $0.023 \pm 0.011 N$, which is less than $1\%$ of the total range of the operating region. Thus, using the model in Section \ref{sec:model}, StRETcH's mechanical response can be estimated given contact conditions and stretch state. 

\section{Experiments}

\begin{figure*}[t]
	\includegraphics[width=\linewidth]{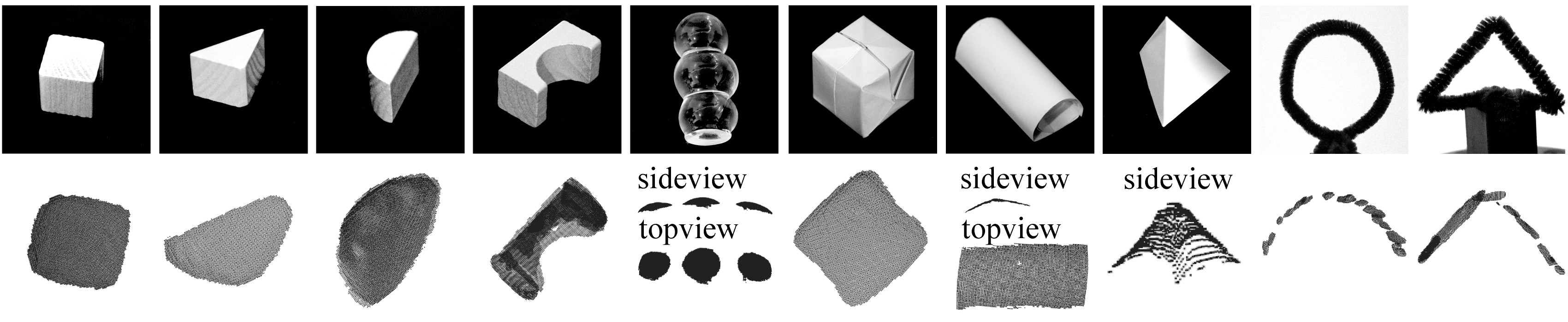}
	\centering
	\caption{3D reconstructions of contacts with a: wooden cube, wooden triangular prism, wooden half-cylinder, wooden bridge, glass snowman, paper cube, paper half-cylinder, paper pyramid, wire circle, and wire triangle.}
	\label{fig:3drendering}
	\vspace{-0pt}
\end{figure*}

\subsection{3D reconstruction of deformable and clear objects}

To demonstrate the capabilities of StRETcH as a tactile hand, we first used it to reconstruct the geometry of several objects. As shown in Figure \ref{fig:3drendering}, StRETcH was capable of reconstructing the contact of several basic geometric shapes as well as soft enough to conform to highly deformable material such as paper and wire. When contact is made, the 3D points associated with the estimated contact area are transformed into the robot world coordinates. As the robot moves to collect a new observation through StRETcH, the object point cloud is updated with every new contact. This allows for the reconstruction of ``multi-view" objects such as the wire circle. For this section, the robot was jogged to multiple positions if necessary (e.g, for the larger objects such as the wooden bridge, snowman, and wire objects). The membrane was kept at its lowest stretch state (zero applied tension) to maintain its high sensitivity and conformability in the case of encountering a deformable object.

The 3D reconstructed geometries match closely to the corresponding objects. From the wooden objects, it is clear that the membrane, at its low stretch state and indentation depth, causes the estimated geometry to be smoother, rounding out the corners and edges of the shapes. However, the dimensions of the contacts are within $2$ mm of error from the real dimensions. The glass snowman, which would normally not be imaged properly through traditional vision, is found to have three domes. The paper 3D objects are imaged with accurate reconstruction, and, despite making contact with StRETcH, the heights of the paper objects are maintained, with an error of $3$ mm. Finally, the robot collected multiple viewpoints of the highly deformable wire circle and triangle, reconstructing the original wire geometry. 

\subsection{Estimation of deformable object stiffness}
\label{sec:estimationstiffness}
\begin{figure}[t]
	\includegraphics[width=\linewidth]{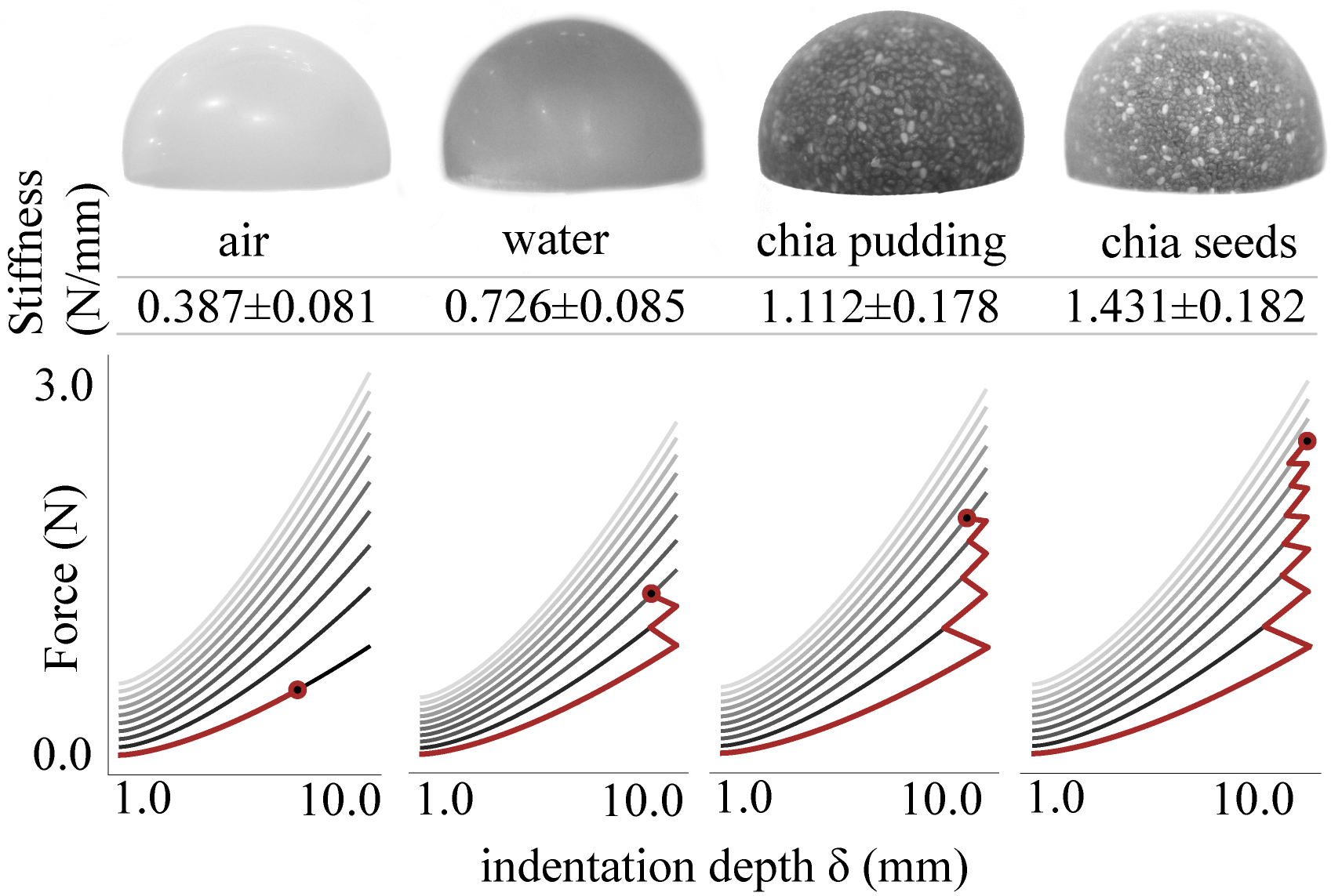}
	\centering
	\caption{(Top): Stiffness estimations in N/mm for balloons filled with air, water, chia pudding, and chia seeds. (Bottom): Active sensing is used to determine stiffness. Plots display force modulation of StRETcH (shown in red) as it increases its indentation depth (moving along the curve to the right) and stretch width (moving to higher curves) until a deformation is sensed. $\delta$-$F$ curves are generated using the learned model described in Section \ref{sec:estimatingloadforce}.}
	\label{fig:stiffnessestimation}
	\vspace{-0pt}
\end{figure}

This section demonstrates how StRETcH actively interacts with a deformable object to estimate its stiffness. Four balloons were filled with approximately the same volume of air, water, a mixture of water and chia seeds, and chia seeds. When water and chia seeds combine, the mixture becomes very viscous relative to water. Qualitatively, the air balloon was the least stiff, followed by water, chia pudding, and chia seeds. Each balloon was fixed inside a cup so it would not move during palpation experiments. The exposed portions of the balloons were approximately hemispherically shaped.

To estimate stiffness, the robot algorithm begins by estimating the radius of the balloon. This is done using the lowest stiffness regime of StRETcH, which, as shown in Section \ref{sec:contactsensitivity}, has greater sensitivity to geometry due to high conformability under zero uniaxial strain. The robot takes note of the balloon's height and presses StRETcH's membrane into the balloon up to 10mm. After observing the contact at 10mm, a sphere is fit to the estimated contact points. Estimating the geometry is necessary, as Figure \ref{fig:sensor_characterization}(e) shows that the effective modulus $E^*$ varies with the contact area. 

The robot proceeds to increase its stretch state incrementally and continues pressing the membrane into the balloon until the height of the balloon is sensed to have dropped below a certain threshold from its original height. We set this to 1.5mm as sensor noise was on the order of $\pm1$mm.  At this point of deformation, the force $F^*$ exerted by the membrane onto the deformable balloon can be calculated using the fitted model described in Section \ref{sec:estimatingloadforce}. 

A proxy for stiffness in N/mm is then determined. Approximating the object as an elastic body with one degree of freedom, the stiffness is calculated by dividing $F^*$ by the deflection of the balloon under the membrane's imposed force. As shown in the upper table of Figure \ref{fig:stiffnessestimation}, StRETcH is capable of repeatably estimating the stiffness of a wide range of substances. Note that, although these are technically proxy values for stiffness due to the simplified representation of the balloons, the estimated values are sufficient for differentiating between materials within this range.  

Finally, Section \ref{sec:estimatingloadforce}'s model can be used to improve the robot's active sensing capabilities. The lower plots of Figure \ref{fig:stiffnessestimation} display example indentation depth ($\delta$) to load force ($F$) curves at different stretch states ($x$), with higher curves corresponding to higher stretch states. The red lines denote the indentation depths and stretch states explored until a deformation is sensed for that particular object. As shown in the plots, if no deformation is sensed by StRETcH before the membrane is stretched any further, the model can be used to calculate the indentation depth to begin palpation for the next stretch state such that the force applied by the membrane is monotonically increasing until the deformation is sensed. This demonstrates that StRETcH can incorporate the model to actively adjust during system identification of the object in question.

\subsection{Rolling a piece of dough and forming a sphere}

\begin{figure}[t]
	\includegraphics[width=\linewidth]{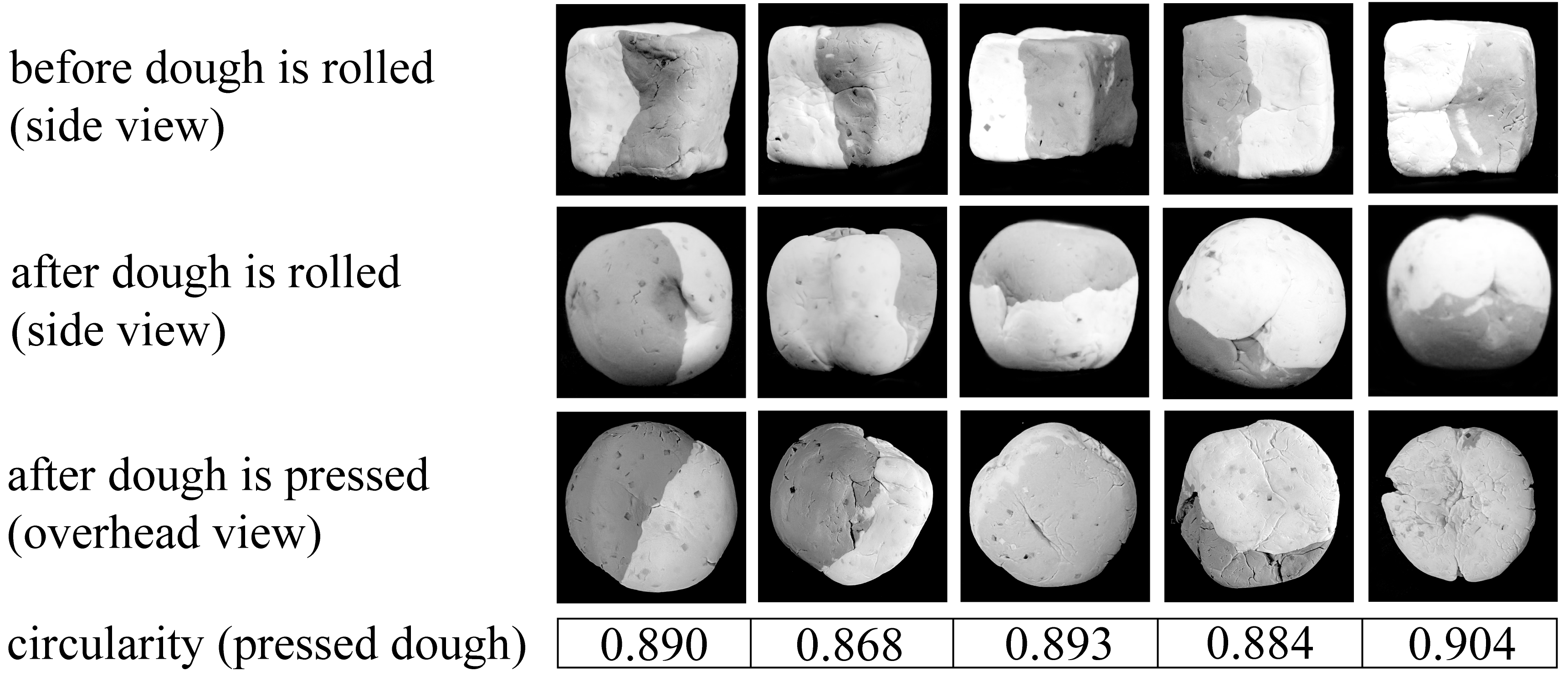}
	\centering
	\caption{Example cookies formed from cubes of Play-Doh. (Top): The dough begins as a cube; (Middle): The dough is rolled into a sphere using medium stiffness; (Bottom): The dough is flattened into a cookie using high stiffness.}
	\label{fig:doughrolling}
	\vspace{-0pt}
\end{figure}

We conclude with a demonstration of the use of StRETcH to form a cookie out of Play-Doh by rolling the dough until it forms a sphere and then pressing it into the shape of a cookie. This task necessitates three different stiffness regimes of StRETcH: (1) low stiffness to measure the geometry of the object; (2) medium stiffness to match the estimated force necessary to deform the object when rolling it; and (3) high stiffness to compress the dough ball into a cookie. 

First, the algorithm presented in Section \ref{sec:estimationstiffness} is used to estimate the force necessary to deform the piece of dough. StRETcH adjusts its stretch state to reflect the estimated stiffness given the current geometry of the dough formation. The UR5 proceeds to then roll the dough in a circle, checking the height of the object in parallel at 1Hz. When the standard deviation of the last ten observed object heights is less than 1mm, the robot stops rolling the piece of dough and assumes it is now a sphere. The high conformability of StRETcH enables tracking of the dough ball's height while providing enough stiffness to deform the dough during rolling. Furthermore, while traction resulting from tangential motions applied to the dough with large indentation depth is not modeled, these additional forces contribute in shaping the dough. Finally, StRETcH engages its highest stiffness by stretching to its maximum width and pressing down onto the dough to form a cookie. If the dough was successfully shaped as a sphere, the cookie would be round. The circularity, or $\frac{4\pi \text{area}}{\text{perimeter}^2}$ of each cookie was calculated, with 0.785 and 1 corresponding to a perfect square and circle, respectively. As shown in Figure \ref{fig:doughrolling}, StRETcH successfully shaped the cookies (with average circularity of 0.888) from cubes of dough.  

\section{Conclusions}

In this work, we presented a Soft to Resistive Elastic Tactile Hand called StRETcH. Modifying the design of the tactile sensor developed in \cite{mcinroe2018towards, huang2019depth}, StRETcH consists of an easily-manufactured membrane that is imaged by a depth sensor and uniaxially stretched by a parallel-jaw gripper. Contact geometry estimation using StRETcH can exhibit as low as sub-millimeter accuracy, and StRETcH can reconstruct the geometry of deformable objects such as paper cubes and wire shapes. The new actuation design enables controllable effective modulus from 4kPa to 9kPa. With a calibrated model of the mechanical response of StRETcH, errors in estimated contact forces are on average less than $\pm0.025$N. Furthermore, by modulating its stiffness, StRETcH was able to efficiently estimate a proxy stiffness for balloons filled with different materials. We concluded with a demonstration of StRETcH working in three different stiffness regimes to 1) estimate the geometry of a piece of dough, 2) deform the dough during rolling, and 3) press the dough ball into a cookie shape. 

This work demonstrates new capabilities of variable stiffness soft tactile hands, and future work aims to improve on the design and modeling of StRETcH as well as to develop more efficient algorithms for manipulating solid deformable objects. First, the stiffness of StRETcH (and therefore the exerted force) is upper-bounded due to the thickness of the membrane and the maximum allowable stretch from the robotic gripper. Future designs of the sensor will focus on exploring new materials that have a larger range of stiffness when stretched (e.g., reinforcing the membrane with flexible fibers) and designing new mechanisms for uniaxial stretching that could miniaturize StRETcH onto an individual robotic finger for more dexterous in-hand manipulation. Furthermore, while the Hertz contact model approximation was shown to fit the experimental data quite closely, we fully acknowledge that a membrane is mechanically quite different and much more complex than an elastic half-space. Therefore, more physically-accurate modeling may be done by pairing state-of-the-art numerical simulation of membranes such as in \cite{taylor2020simulation} with a learning-based approach presented in \cite{narang2020interpreting} to estimate the contact force and stiffness of an object from contact geometry and the stretch state of the sensor. One could also imagine using a highly stretchable sensing skin such as in \cite{larson2016highly} to directly sense the internal strain of the membrane, which, paired with numerical simulation, could potentially make more accurate predictions of contact forces. Finally, StRETcH was able to roll a block of dough into roughly a sphere, but each roll-out required on average ten minutes, with some lasting up to twenty. Future work explores using reinforcement learning techniques with StRETcH to efficiently shape dough into different desired geometries. 

{\footnotesize
\section*{Acknowledgment} The authors were supported in part by the National Science Foundation Graduate Research Fellowship. We thank Matthew Matl, Isabella Huang, Yashraj Narang, David Steigmann, Milad Shirani, and Michael Taylor for their insightful feedback and suggestions.
}



\begin{thebibliography}{46}
\bibitem{sanchez2018robotic}
J. Sanchez, J.-A. Corrales, B.-C. Bouzgarrou, and Y. Mezouar,
"Robotic manipulation and sensing of deformable objects
in domestic and industrial applications: A survey", {\em The International Journal of Robotics Research}, vol. 37, no. 7, pp. 688–716, 2018.
  
\bibitem{yousef2011tactile}
H. Yousef, M. Boukallel, and K. Althoefer, “Tactile sensing
for dexterous in-hand manipulation in robotics—a review”,
{\em Sensors and Actuators A: physical}, vol. 167, no. 2, pp. 171–187, 2011.

\bibitem{luo2017robotic}
S. Luo, J. Bimbo, R. Dahiya, and H. Liu, “Robotic tactile
perception of object properties: A review”, {\em Mechatronics},
vol. 48, pp. 54–67, 2017.

\bibitem{ward2018tactip}
B. Ward-Cherrier, N. Pestell, L. Cramphorn, B. Winstone,
M. E. Giannaccini, J. Rossiter, and N. F. Lepora, “The tactip
family: Soft optical tactile sensors with 3d-printed biomimetic
morphologies”, {\em Soft robotics}, vol. 5, no. 2, pp. 216–227, 2018.

\bibitem{yuan2018active}
W. Yuan, Y. Mo, S. Wang, and E. H. Adelson, “Active
clothing material perception using tactile sensing and deep
learning”, in {\em 2018 IEEE International Conference on Robotics
and Automation (ICRA)}, IEEE, 2018, pp. 1–8.

\bibitem{kuppuswamy2020soft}
N. Kuppuswamy, A. Alspach, A. Uttamchandani, S. Creasey, T. Ikeda and R. Tedrake, "Soft-bubble grippers for robust and perceptive manipulation," in {\em 2020 IEEE/RSJ International Conference on Intelligent Robots and Systems (IROS)}, 2020, pp. 9917-9924.

\bibitem{mcinroe2018towards}
B. W. McInroe, C. L. Chen, K. Y. Goldberg, R. Bajcsy,
and R. S. Fearing, “Towards a soft fingertip with integrated
sensing and actuation”, in {\em 2018 IEEE/RSJ International
Conference on Intelligent Robots and Systems (IROS)}, IEEE,
2018, pp. 6437–6444.

\bibitem{huang2019depth} 
I. Huang, J. Liu, and R. Bajcsy, “A depth camera-based
soft fingertip device for contact region estimation and
perception-action coupling”, in {\em 2019 International Conference
on Robotics and Automation (ICRA)}, IEEE, 2019, pp. 8443–
8449.

\bibitem{huang2020high}
I. Huang and R. Bajcsy, “High resolution soft tactile interface
for physical human-robot interaction”, in {\em 2020 IEEE International
Conference on Robotics and Automation (ICRA)},
IEEE, 2020, pp. 1705–1711.

\bibitem{huang2020robot}
——, “Robot learning from demonstration with tactile
signals for geometry-dependent tasks”, in {\em 2020 IEEE/RSJ
International Conference on Intelligent Robots and Systems
(IROS)}, IEEE, 2020.

\bibitem{hammond2012soft}
F. L. Hammond, R. K. Kramer, Q. Wan, R. D. Howe, and
R. J. Wood, “Soft tactile sensor arrays for micromanipulation”,
in {\em 2012 IEEE/RSJ International Conference on Intelligent
Robots and Systems}, IEEE, 2012, pp. 25–32.

\bibitem{lee2019neuro}
W. W. Lee, Y. J. Tan, H. Yao, S. Li, H. H. See, M. Hon,
K. A. Ng, B. Xiong, J. S. Ho, and B. C. Tee, “A neuro-inspired
artificial peripheral nervous system for scalable electronic
skins”, {\em Science Robotics}, vol. 4, no. 32, eaax2198, 2019.

\bibitem{suh2014soft}
C. Suh, J. C. Margarit, Y. S. Song, and J. Paik, “Soft
pneumatic actuator skin with embedded sensors”, in {\em 2014
IEEE/RSJ International Conference on Intelligent Robots and
Systems}, Ieee, 2014, pp. 2783–2788.

\bibitem{drimus2011classification}
A. Drimus, G. Kootstra, A. Bilberg, and D. Kragic, “Classification
of rigid and deformable objects using a novel tactile
sensor”, in {\em 2011 15th International Conference on Advanced
Robotics (ICAR)}, IEEE, 2011, pp. 427–434.

\bibitem{buscher2015flexible}
G. H. Buscher, R. Koiva, C. Schurmann, R. Haschke, and H. J.
Ritter, “Flexible and stretchable fabric-based tactile sensor”,
{\em Robotics and Autonomous Systems}, vol. 63, pp. 244–252,
2015.

\bibitem{atalay2017batch}
A. Atalay, V. Sanchez, O. Atalay, D. M. Vogt, F. Haufe, R. J.
Wood, and C. J. Walsh, “Batch fabrication of customizable
silicone-textile composite capacitive strain sensors for human motion tracking”, {\em Advanced Materials Technologies}, vol. 2,
no. 9, p. 1 700 136, 2017.

\bibitem{yoshikai2009development}
T. Yoshikai, H. Fukushima, M. Hayashi, and M. Inaba, “Development
of soft stretchable knit sensor for humanoids’ wholebody
tactile sensibility”, in {\em 2009 9th IEEE-RAS International
Conference on Humanoid Robots}, IEEE, 2009, pp. 624–631.

\bibitem{larson2016highly}
C. Larson, B. Peele, S. Li, S. Robinson, M. Totaro, L.
Beccai, B. Mazzolai, and R. Shepherd, “Highly stretchable
electroluminescent skin for optical signaling and tactile
sensing”, {\em Science}, vol. 351, no. 6277, pp. 1071–1074, 2016.

\bibitem{pu2017ultrastretchable}
X. Pu, M. Liu, X. Chen, J. Sun, C. Du, Y. Zhang, J. Zhai,
W. Hu, and Z. L. Wang, “Ultrastretchable, transparent triboelectric
nanogenerator as electronic skin for biomechanical
energy harvesting and tactile sensing”, {\em Science advances},
vol. 3, no. 5, e1700015, 2017.

\bibitem{narang2020interpreting}
Y. S. Narang, K. Van Wyk, A. Mousavian, and D. Fox,  "Interpreting and Predicting Tactile Signals via a Physics-Based and Data-Driven Framework", {\em RSS}, vol. 16, no. 84, 2020.  

\bibitem{yamaguchi2016combining}
A. Yamaguchi and C. G. Atkeson, “Combining finger
vision and optical tactile sensing: Reducing and handling
errors while cutting vegetables”, in {\em 2016 IEEE-RAS 16th
International Conference on Humanoid Robots (Humanoids)},
IEEE, 2016, pp. 1045–1051.

\bibitem{yuan2017shape}
W. Yuan, C. Zhu, A. Owens, M. A. Srinivasan, and E. H.
Adelson, “Shape-independent hardness estimation using deep
learning and a gelsight tactile sensor”, in {\em 2017 IEEE International
Conference on Robotics and Automation (ICRA)},
IEEE, 2017, pp. 951–958.

\bibitem{alspach2019soft}
A. Alspach, K. Hashimoto, N. Kuppuswarny, and R. Tedrake,
“Soft-bubble: A highly compliant dense geometry tactile
sensor for robot manipulation”, in {\em 2019 2nd IEEE International
Conference on Soft Robotics (RoboSoft)}, IEEE, 2019,
pp. 597–604.

\bibitem{shikida2003active}
M. Shikida, T. Shimizu, K. Sato, and K. Itoigawa, “Active
tactile sensor for detecting contact force and hardness of an
object”, {\em Sensors and Actuators A: physical}, vol. 103, no. 1-2,
pp. 213–218, 2003.

\bibitem{yussof2008determination}
H. Yussof, M. Ohka, A. R. Omar, and M. A. Ayub, “Determination
of object stiffness control parameters in robot
manipulation using a prototype optical three-axis tactile
sensor”, in {\em SENSORS, 2008 IEEE}, IEEE, 2008, pp. 992–995.

\bibitem{ju2018variable}
F. Ju, Y. Yun, Z. Zhang, Y. Wang, Y. Wang, L. Zhang, and
B. Chen, “A variable-impedance piezoelectric tactile sensor
with tunable sensing performance for tissue hardness sensing
in robotic tumor palpation”, {\em Smart Materials and Structures},
vol. 27, no. 11, p. 115 039, 2018.

\bibitem{omata2004real}
S. Omata, Y. Murayama, and C. E. Constantinou, “Real time
robotic tactile sensor system for the determination of the
physical properties of biomaterials”, {\em Sensors and Actuators
A: Physical}, vol. 112, no. 2-3, pp. 278–285, 2004.

\bibitem{yan2020learning}
W. Yan, A. Vangipuram, P. Abbeel, and L. Pinto, “Learning
predictive representations for deformable objects using contrastive
estimation”, {\em Conference on Robot Learning}, 2020.

\bibitem{sundaresan2020learning}
Sundaresan, Priya, et al. "Learning rope manipulation policies using dense object descriptors trained on synthetic depth data." {\em 2020 IEEE International Conference on Robotics and Automation (ICRA)}. IEEE, 2020.

\bibitem{balkcom2008robotic}
D. J. Balkcom and M. T. Mason, “Robotic origami folding”,
{\em The International Journal of Robotics Research}, vol. 27, no. 5,
pp. 613–627, 2008.

\bibitem{stria2014garment}
J. Stria, D. Prša, V. Hlaváˇc, L. Wagner, V. Petrık, P. Krsek,
and V. Smutn, “Garment perception and its folding using a
dual-arm robot”, in {\em 2014 IEEE/RSJ International Conference
on Intelligent Robots and Systems}, IEEE, 2014, pp. 61–67.

\bibitem{miller2012geometric}
S. Miller, J. Van Den Berg, M. Fritz, T. Darrell, K. Goldberg,
and P. Abbeel, “A geometric approach to robotic laundry folding”, {\em The International Journal of Robotics Research},
vol. 31, no. 2, pp. 249–267, 2012.

\bibitem{navarro2016automatic}
D. Navarro-Alarcon, H. M. Yip, Z. Wang, Y.-H. Liu, F. Zhong,
T. Zhang, and P. Li, “Automatic 3-d manipulation of soft
objects by robotic arms with an adaptive deformation model”,
{\em IEEE Transactions on Robotics}, vol. 32, no. 2, pp. 429–441,
2016.

\bibitem{navarro2017fourier}
D. Navarro-Alarcon and Y.-H. Liu, “Fourier-based shape
servoing: A new feedback method to actively deform soft
objects into desired 2-d image contours”, {\em IEEE Transactions
on Robotics}, vol. 34, no. 1, pp. 272–279, 2017.

\bibitem{shah2019morphing}
D. S. Shah, M. C. Yuen, L. G. Tilton, E. J. Yang, and R.
Kramer-Bottiglio, “Morphing robots using robotic skins that
sculpt clay”, {\em IEEE Robotics and Automation Letters}, vol. 4,
no. 2, pp. 2204–2211, 2019.

\bibitem{figueroa2016learning}
N. Figueroa, A. L. P. Ureche, and A. Billard, “Learning
complex sequential tasks from demonstration: A pizza dough
rolling case study”, in {\em 2016 11th ACM/IEEE International
Conference on Human-Robot Interaction (HRI)}, Ieee, 2016,
pp. 611–612.

\bibitem{nascimento2012brand}
E. R. Nascimento, G. L. Oliveira, M. F. Campos, A. W. Vieira,
and W. R. Schwartz, “Brand: A robust appearance and depth
descriptor for rgb-d images”, in {\em 2012 IEEE/RSJ International
Conference on Intelligent Robots and Systems}, IEEE, 2012,
pp. 1720–1726.

\bibitem{bhatia1968finite}
N. M. Bhatia and W. Nachbar, “Finite indentation of an elastic
membrane by a spherical indenter”, {\em International Journal of
Non-Linear Mechanics}, vol. 3, no. 3, pp. 307–324, 1968.

\bibitem{yang1971indentation}
W. Yang and K. Hsu, “Indentation of a circular membrane”,
1971.

\bibitem{begley2004spherical}
M. R. Begley and T. J. Mackin, “Spherical indentation of
freestanding circular thin films in the membrane regime”,
{\em Journal of the Mechanics and Physics of Solids}, vol. 52,
no. 9, pp. 2005–2023, 2004.

\bibitem{pamplona2014analytical}
D. C. Pamplona, H. I. Weber, and G. R. Sampaio, “Analytical,
numerical and experimental analysis of continuous indentation
of a flat hyperelastic circular membrane by a rigid cylindrical
indenter”, {\em International Journal of Mechanical Sciences},
vol. 87, pp. 18–25, 2014.

\bibitem{liu2018puncture}
J. Liu, Z. Chen, X. Liang, X. Huang, G. Mao, W. Hong, H. Yu,
and S. Qu, “Puncture mechanics of soft elastomeric membrane
with large deformation by rigid cylindrical indenter”, {\em Journal
of the Mechanics and Physics of Solids}, vol. 112, pp. 458–471,
2018.

\bibitem{taylor2019finite}
M. Taylor, M. Shirani, Y. Dabiri, J. Guccione, and D.
Steigmann, “Finite elastic wrinkling deformations of incompressible
fiber-reinforced plates”, {\em International Journal of
Engineering Science}, vol. 144, p. 103 138, 2019.

\bibitem{scott2004indentation}
O. Scott, M. Begley, U. Komaragiri, and T. Mackin, “Indentation
of freestanding circular elastomer films using spherical
indenters”, {\em Acta materialia}, vol. 52, no. 16, pp. 4877–4885,
2004.

\bibitem{johnson1987contact}
K. L. Johnson and K. L. Johnson, {\em Contact mechanics}.
Cambridge university press, 1987.

\bibitem{taylor2020simulation}
M. Taylor and M. Shirani, “Simulation of wrinkling in
incompressible anisotropic thin sheets with wavy fibers”,
{\em International Journal of Non-Linear Mechanics}, p. 103 610,
2020.

\end{thebibliography}
\end{document}